\documentclass[11pt, a4paper, logo, copyright]{googledeepmind}

\usepackage[authoryear, sort&compress, round]{natbib}
\usepackage{times}
\usepackage{latexsym}
\usepackage{lineno}

\usepackage[utf8]{inputenc} 
\usepackage[T1]{fontenc}    
\usepackage{hyperref}       
\usepackage{url}            
\usepackage{booktabs}       
\usepackage{amsfonts}       
\usepackage{nicefrac}       
\usepackage{microtype}      
\usepackage[table,dvipsnames]{xcolor}         
\usepackage{xspace}
\usepackage{pifont}
\usepackage{booktabs}
\usepackage{multirow}
\usepackage{graphicx}
\usepackage{wrapfig}
\usepackage{makecell}
\usepackage{xspace}
\usepackage{comment}
\usepackage[disable]{todonotes}
\usepackage{enumitem}
\usepackage{amsmath}
\usepackage{titlesec}
\usepackage[most]{tcolorbox}

\usepackage{tcolorbox}
\newtcolorbox{prompt}[1]{colback=gray!20,colframe=gray!50!black,fonttitle=\bfseries,title=#1}

\title{CoReflect: A Reflective Co-Evolution Framework for Improving Conversational Evaluation}

\correspondingauthor{yunzhel2@illinois.edu}


\reportnumber{} 
\author[$\dagger$,1]{Yunzhe Li}
\author[2]{Richie Yueqi Feng}
\author[$\dagger$,1]{Tianxin Wei}
\author[2]{Chin-Chia Hsu}
\affil[$\dagger$]{Work done while at Google DeepMind}
\affil[1]{University of Illinois Urbana-Champaign}
\affil[2]{Google DeepMind}

\begin{abstract}
Evaluating conversational systems in multi-turn settings remains a fundamental challenge. 
Conventional pipelines typically rely on manually defined rubrics and fixed conversational context\textemdash a static approach that limits coverage and fails to capture the diverse, emergent behaviors of dialogue models.
To address this, we introduce CoReflect (A Reflective Co-Evolution Framework for Improving Conversational Evaluation), which unifies dialogue simulation and evaluation into an adaptive, iterative process. 
CoReflect employs a conversation planner that generates structured templates to guide a user simulator through diverse, goal-directed dialogues. 
Subsequently, a reflective analyzer processes these dialogues to identify systematic behavioral patterns and automatically refine the evaluation rubrics.
Crucially, the insights from the conversation analysis are fed back into the planner to update conversation templates for subsequent iterations. This co-evolution loop ensures that the complexity of test cases and the diagnostic precision of rubrics improve in tandem. 
By minimizing human intervention, CoReflect provides a scalable and self-refining methodology that allows evaluation protocols to adapt alongside the rapidly advancing capabilities of dialogue models.

\end{abstract}

\begin{document}

\maketitle

\addtocontents{toc}{\protect\setcounter{tocdepth}{-1}}

\section{Introduction}

Large language models (LLMs) increasingly power multi-turn applications such as personalized tutors
and customer service agents \citep{brown2020language,openai2023gpt4}.
In these dynamic settings, users demand more than linguistic fluency; they expect coherent, adaptive,
and contextually grounded dialogues. Evaluating such  conversational capabilities, however, remains a challenging open problem.

To date, the most reliable approach to conversational evaluation relies on open-ended human assessment against predefined rubrics, as human raters can dynamically probe nuanced behaviors and value alignment over extended interactions. However, this method is costly, time-consuming, and prone to subjective bias. While automated alternatives, such as ``LLM-as-a-judge'' frameworks, offer efficiency and have shown 
promising alignment with human judgments \citep{liu2023geval,zheng2023mtbench, zhao2025prefeval}, they often rely on
static, pre-designed conversational contexts that fail to capture authentic model behavior.
Furthermore, LLM-based simulators, while capable of producing fluent and natural conversational interactions, often struggle to systematically elicit challenging or adversarial behaviors, and may exhibit shallow exploration or mode collapse in self-driven interactions \citep{park2023generative,shanahan2023roleplay,chiang2024chatbotarena, wei2025evo}.

More fundamentally, both human and automated evaluation approaches can be constrained by the evaluation architect's subjective focus and limited foresight. Overly narrow rubrics may fail to capture emergent model capabilities or unforeseen behaviors that manifest during subsequent deployment, leading to costly redesign and re-execution of the evaluation process.



To address this limitation, we propose \textbf{CoReflect}, a co-evolutionary and reflective evaluation framework that moves beyond static benchmarks and fixed rubrics. By introducing an adaptive, iterative process, CoReflect ensures that both dialogue simulation and evaluation criteria evolve in tandem to capture model behaviors. Instead of relying on static conversational scripts, the framework utilizes a \textit{conversation planner} as the central engine that orchestrates the entire interaction under specific user persona and scenario. Specifically, the planner generates structured, goal-directed templates that prescribe specific turn-level instructions for a user simulator on how to interact with test models. 

Given these planner-guided interactions with the test models, an \textit{LLM-as-a-judge evaluator} then assesses the resulting conversations using a structured three-step protocol---analyzing turns, synthesizing observations, and assigning ratings against the current rubrics. 
Following the evaluation results, a \textit{reflective analyzer} processes these conversation assessments to cluster behavioral patterns and systematic failures, extracting insights to refine the evaluation rubrics automatically.
These insights are also fed back to the planner, which updates its conversation templates to target identified weaknesses with greater precision in subsequent iterations.



Our main contributions are presented  as follows:
\begin{itemize}[leftmargin=*]

\item \textbf{Co-evolutionary conversational evaluation framework.} 
We propose CoReflect, a novel co-evolutionary evaluation framework in which conversation templates and evaluation rubrics are jointly refined through an iterative feedback loop. This co-evolution mechanism is the core innovation of our work, allowing the user simulation and the diagnostic precision of the rating criteria improve in tandem.

\item \textbf{Autonomous synthesis of evaluation rubrics.} 
CoReflect introduces a reflective analyzer as the primary mechanism for translating model behavioral data into structured evaluation logic, effectively alleviating the human burden of manual design. By autonomously synthesizing patterns from simulated interactions, it mitigates subjective human bias and enables the system to dynamically expand its scope beyond static rubrics. This data-driven approach allows CoReflect to capture unexpected model behaviors that might otherwise remain hidden, ensuring the evaluation framework evolves in lockstep with the model's complexity.


\item \textbf{Comprehensive empirical validation.} 
We conduct extensive automatic and human-in-the-loop experiments across diverse personas, scenarios, and model families, demonstrating the ability of CoReflect to support comprehensive and behavior-sensitive evaluation of state-of-the-art LLMs in multi-turn conversations.
\end{itemize}

\section{Methodology}

\begin{figure*}[htb]
    \centering
    \includegraphics[width=1\linewidth]{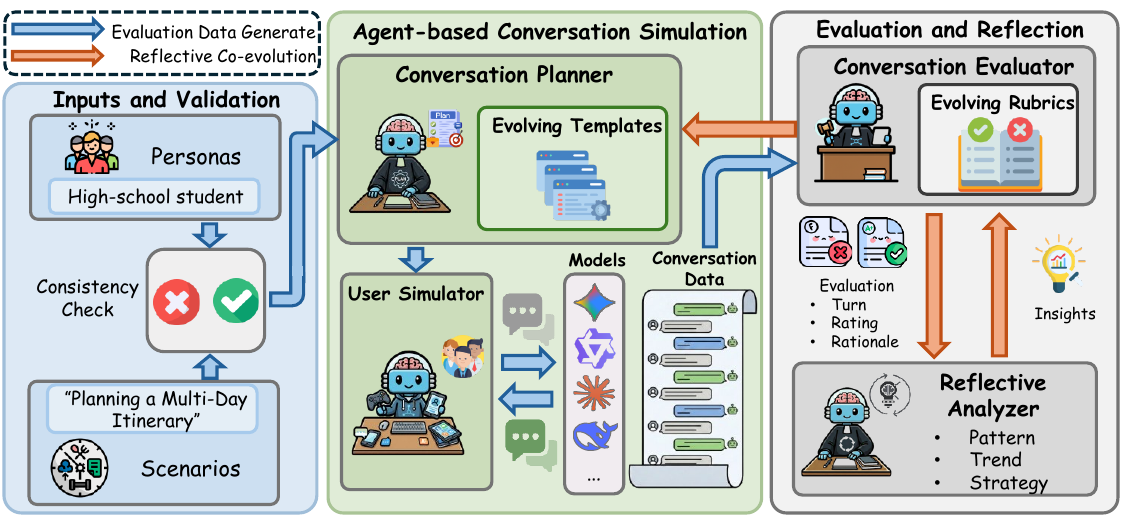}
\caption{
Overview of the CoReflect framework. 
Initially, evaluation instances are generated from persona-scenario pairs and validated through a consistency check to ensure contextual coherence. Then the core co-evolution loop proceeds through two parts: 
(i) Conversation simulation, where a planner generates structured conversation templates to guide a user simulator through goal-directed dialogues across diverse persona-scenario pairs; and (ii) Reflective rubric refinement, where an LLM-as-a-judge evaluates simulated interactions and a reflective analyzer extracts insights from clustered behavioral patterns to refine evaluation rubrics. 
The insights are fed back to the planner to update conversation templates for subsequent iterations. 
}

    \label{fig:framework}
\end{figure*}

The evaluation process is initialized by a human architect who manually defines a set of coarse, high-level rubrics to serve as the foundational criteria for assessing the conversational quality of the test models.
In each iteration, each model interacts with a user simulator across various persona–scenario pairs, guided by \emph{a conversation planner}. This planner generates goal-directed, persona-grounded conversation templates that instructs the user simulator on how to conduct each user turn, thereby precisely challenging the models against the given rubrics.

Following these interactions, \emph{an evaluator} assesses the conversations based on the established rubrics, providing quantitative ratings and qualitative rationales.
These results are then sampled and processed by \emph{a reflective analyzer} to characterize model behaviors ad limitations, offering insights that drive an iterative refinement loop; in this loop, the rubrics are tuned and the planner accordingly updates its templates to better challenge the models in subsequent rounds.
Figure~\ref{fig:framework} illustrates this interplay between the planner, evaluator, and reflective analyzer.
This process proceeds through time steps $t = 1, 2,\dots T$, where $t=1$ represents the initial state, and terminates after $T$ iterations.

We provide the specifics of the framework and the process as follows.

\subsection{Personas, Scenarios, and Test Models}
Each evaluation \emph{instance} in this study comprises two components: (i) a user, who is manifested through a persona profile, (ii) a scenario, which defines a context, task, and objective of the conversation.
Every test model is evaluated within these personas and scenarios pairings, with the goal of fulfilling the user’s intent throughout the conversation. 
Below, we elaborate the construction and the pairing process of personas and scenarios to generate the evaluation set.

\paragraph{Personas.}
To capture user behaviors in our evaluation, we create structured personas that specify how users express themselves and how they expect the system to interact with them.
Specifically, each persona comprises two parts:
(1) \textbf{User traits}: the user's attributes manifested in their communication, such as tone, verbosity, and conversational nuances. We denote user traits by $x$. 
(2) \textbf{Response preferences}: the desired qualities of the model's responses, including reasoning depth, level of detail, and formality, etc. We denote user preferences by $\theta$. 
This representation enables controlled variation across communication styles while maintaining realism and interpretability.  
Complete schema definitions and examples are provided in Appendix~A.

\begin{table*}[ht]
\centering
\small
\begin{tabular}{@{}p{2cm}@{\hspace{5pt}}p{3cm}@{\hspace{5pt}}p{2.6cm}@{\hspace{5pt}}p{3cm}@{\hspace{5pt}}p{4cm}@{}}
\toprule
\textbf{Category} & \textbf{User Intent} & \textbf{Assistant Role} & \textbf{Typical Outputs} & \textbf{Example} \\
\midrule
Instructional & Acquire knowledge or develop skills & Tutor / Instructor & Explanations, practice questions, scaffolds & \emph{``Teach me how neural networks work.''} \\
\addlinespace
Informational  & Retrieve facts or clarify references & Reference / Expert & Factual answers, summaries, citations & \emph{``What is the GDP of Brazil in 2023?''} \\
\addlinespace
Operational  & Complete a concrete task or create content & Productivity Tool & Code, documents, plans, translations & \emph{``Write a weekly report summary for my team.''} \\
\addlinespace
Interactive  & Engage socially, emotionally, or creatively & Companion / Partner & Stories, dialogues, empathetic replies & \emph{``I feel nervous about my exam---can we talk about it?''} \\
\bottomrule
\end{tabular}
\caption{Four scenario categories  in the evaluation dataset $\mathcal{D}$, defined by specific user intents, assigned assistant roles, characteristic outputs, and representative examples.}
\label{tab:scenario_categories}
\vspace{-6pt}
\end{table*}

\paragraph{Scenarios.}
Scenarios are created independently from personas to ensure broad coverage of conversational goals. 
Each scenario specifies a context, primary task, objective of the conversation, and an estimate of number of turns to complete the task.
We categorize scenarios into four intent types--\emph{instructional}, \emph{informational}, \emph{operational}, and \emph{interactive}--as detailed in Table~\ref{tab:scenario_categories}.
Collectively, these categories encompass factual, task-oriented, and social interactions.
We denote a scenario by $s$.

\paragraph{Pairing process and consistency check.}
Each persona is paired with every scenario. 
However, not every persona–scenario combination yields a contextually-coherent setup (e.g., a young student tutoring a senior engineer). 
To ensure plausibility, each persona–scenario pair undergoes a consistency check conducted by a model-based verifier to assess the contextual coherence in each pair. 
Only those pairs validated as plausible are retained for simulation and evaluation.  
Formally, let $\mathcal{D}$ denote this synthetic dataset of validated persona–scenario pairs. We represent the $i$-th pair as a tuple $(x_i, \theta_i, s_i)$.
More details on consistency check can be found in Appendix~C.

\paragraph{Test models.} Each test model is evaluated against the whole set $\mathcal{D}$.
The collection of test models is denoted by $\mathcal{M} \triangleq \{m_j\}_{j=1}^M$, where each $m_j$ represents a specific conversational model and $M=|\mathcal{M}|$ is the total number of the models.

\subsection{Planner and Conversation Simulation}
Let $K$ denote the number of rubrics considered in the evaluation. We define the set of rubrics at the start of iteration $t$ by $\mathcal{R}^{(t)}\triangleq \{R_k^{(t)}\}_{k=1}^{K}$.
Whereas the initial set $\mathcal{R}^{(1)}$ is manually drafted, these rubrics are  refined in subsequent iterations.

To improve template quality, incorporates a set of insights from the preceding iteration, denoted by $\mathcal{I}^{(t-1)}$.
We initialize $\mathcal{I}^{(0)}=\emptyset$, and will explain how insights are derived in Section 2.4.

At iteration $t$ with insights $\mathcal{I}^{(t-1)}$, for each validated instance $(x_i, \theta_i, s_i)\in \mathcal{D}$, the planner determines the number of turns $N_i^{(t)}$ required to complete the task and generates a structured conversation template $\mathcal{T}_i^{(t)}$ that encodes specific instructions for every user turn.  
These turn-level instructions guide the dialogue, allowing the conversation to unfold naturally while systematically probing specific model capabilities (e.g., reasoning, clarification). 
Appendix~D describes the template schema.

Once the templates for all persona–scenario pairs in $\mathcal{D}$ are created, each model $m_j \in \mathcal{M}$ conducts a dialogue with the user simulator for every pair.
The simulator is provided with the instance $(x_i, \theta_i, s_i)$ and the corresponding template $\mathcal{A}_i^{(t)}$, which serve as contextual grounding and turn-level instructions.
In contrast, model $m_j$ has only user traits $x_i$ as input information.
The simulator and model $m_j$ then have a dialogue, based on their respective information sets and the evolving conversation history.
This asymmetric information is designed to simulate real-world chatbots leveraging historical personal data (i.e., $x_i$) to personalize interactions.
We denote the resulting conversation generated by model $m_j$ for the $i$-th instance at iteration $t$ as $C_{i,j}^{(t)}$.


\subsection{Conversation Evaluator}
An LLM-as-a-judge evaluates each generated conversation $C_{i,j}^{(t)}$ by model $m_j$ against the corresponding instance $(x_i, \theta_i, s_i)$ and the rubric set $\mathcal{R}^{(t)}$. 

To ensure evaluative reliability and interpretability, the judge generates a rating and a corresponding rationale for each rubric, following a structured three-step reasoning protocol.
First, it performs a turn-level analysis to generate concise observations for every model response grounded in the immediate context. Next, it aggregates these turn-level  observations to identify conversation-wide strengths and weaknesses. Finally, the evaluator determines a final rating, assigning a numerical score to each rubric alongside a rationale that references the synthesized evidence.

Let $r_{i,j,k}^{(t)}$ and $\rho_{i,j,k}^{(t)}$ denote the rating and rationale for conversation $C_{i,j}^{(t)}$ as evaluated under rubric $R_k^{(t)} \in \mathcal{R}^{(t)}$. The performance of model $m_j$ for a specific rubric $R_k^{(t)}$ is defined as the mean rating across the conversation set $\mathcal{D}$; the model's aggregate performance, $\mu_j^{(t)}$, is then calculated as the average across all $K$ rubrics:
{\small
\setlength{\abovedisplayskip}{4pt}
\setlength{\belowdisplayskip}{4pt}
\setlength{\abovedisplayshortskip}{2pt}
\setlength{\belowdisplayshortskip}{2pt}
\begin{align*}
&\mu_{j,k}^{(t)} \triangleq \frac{1}{|\mathcal{D}|}\sum_{i=1}^{|\mathcal{D}|} r_{i,j,k}^{(t)}, \quad k=1,\dots,K \\
&\mu_j^{(t)} \triangleq \frac{1}{K}\sum_{k=1}^{K}\mu_{j,k}^{(t)} .
\end{align*}
}
To measure  performance consistency, we first define the conversation-level rating $\tilde{r}_{i,j}^{(t)}$ for each $C_{i,j}$ as the average rating across all $K$ rubrics, i.e., $\tilde{r}_{i,j}^{(t)} \triangleq \frac{1}{K}\sum_{k=1}^K r_{i,j,k}^{(t)}.$ We then compute the variance of these ratings across the conversations $\mathcal{D}$:
{\small
\setlength{\abovedisplayskip}{4pt}
\setlength{\belowdisplayskip}{4pt}
\begin{align*}
(\sigma_{j}^{(t)})^2 \triangleq 
\frac{1}{|\mathcal{D}| - 1} 
\sum_{i=1}^{|\mathcal{D}|} 
\bigl(\tilde{r}_{i,j}^{(t)} - \mu_j^{(t)}\bigr)^2 .
\end{align*}
}
This quantifies the stability of model $m_j$ across diverse scenarios, where a lower variance indicates more predictable performance.


\subsection{Reflective Analyzer for Rubric Refinement and Planner Feedback}



A reflective analyzer synthesizes the collected ratings $\{r_{i,j,k}^{(t)}\}$ and rationales $\{\rho_{i,j,k}^{(t)}\}$ to extract insights $\mathcal{I}^{(t)}$ regarding model behavior. These insights are then leveraged to sharpen the rubrics' diagnostic precision and to refine the planner's templates, enabling a more effective testing of model strengths and weakness in subsequent iterations.
The reflective analyzer proceeds as follows:

\begin{enumerate}[label=(\arabic*)]
    \item \textbf{Instance sampling:}    
    Based on the assigned ratings $\{r_{i,j,k}^{(t)}\}$, we partition the conversations into two subsets representing high- and low-rated tiers.  
    To ensure a balanced and computationally efficient analysis, we draw an equal number of samples from each tier to construct an analysis pool.
    For each sampled conversation, we retain only the associated rationale $\rho_{i,j,k}^{(t)}$ as the structured evidence for subsequent behavioral analysis.

    \item \textbf{Pattern discovery:}  
    The sampled rationales are embedded and clustered into a set of \emph{behavioral families}, $\mathcal{F}^{(t)} \triangleq \{f_\ell^{(t)}\}_{\ell=1}^{L^{(t)}}$, where $L^{(t)}$ represents the number of distinct, recurring patterns identified at iteration $t$.
    Each family $f_\ell$ characterizes a specific conversational pattern, including both undesirable behaviors
    (e.g., loss of task focus, contradiction to prior context, stylistic drift)
    and effective strategies
    (e.g., proactive clarification, structured task decomposition). 

    \item \textbf{Insight synthesis and rubric refinement.}  
    For every family $f_\ell^{(t)}$, the reflective analyzer produces an interpretable insight $\iota_\ell^{(t)}$ that characterizes the underlying model behavior and specifies the criteria for reward or penalty. 
    These individual insights are aggregated into a comprehensive set $\mathcal{I}^{(t)} \triangleq \{\iota_\ell^{(t)}\}_{\ell=1}^{L^{(t)}}$.
    These insights are utilized to revise the corresponding rubrics in $\mathcal{R}^{(t)}$, specifically by updating performance definitions, rating anchors, and evidence cues. This process enhances the framework's diagnostic precision for the subsequent iteration, formalized as:
    
{\small
\setlength{\abovedisplayskip}{4pt}
\setlength{\belowdisplayskip}{4pt}
\[
\mathcal{R}^{(t+1)} \leftarrow 
\text{UPDATE}\!\left(\mathcal{R}^{(t)},\, \mathcal{I}^{(t)}\right)
\]w
}
    As previously noted, the derived insights $\mathcal{I}^{(t)}$ are inputs for the conversation planner. This integration refines the conversation templates to target identified behaviors with better precision in the subsequent iteration.
\end{enumerate}

\subsection{Measures on Rubrics Refinement}

To assess whether iterative rubric refinement improves evaluation quality, we track two complementary measures across iterations $t$: rubric \emph{discriminability} and \emph{stability}. 
Rubric discriminability, denoted by $\Delta^{(t)}$, captures the degree to which a rubric separates models with different capability levels, operationalized as the inter-model variability of mean ratings.
Rubric stability is measured by $\Gamma^{(t)}$, which evaluates rating consistency through intra-model variance, together with rank consistency measured by Spearman’s $\rho$.
Their formal definitions are provided in Appendix~\ref{app:rubric_metrics}.

\section{Experiments}\label{sec:exp}

\subsection{Experiment Setup}
\paragraph{Datasets.} We created $30$ personas, each embodying distinct communication styles and response preferences, along with $10$ scenarios per category (cf. Table~\ref{tab:scenario_categories}), yielding $953$ validated persona-scenario pairs $\mathcal{D}$. after automated consistency checks.
Summary statistics are provided in Table~\ref{tab:conversation_data}.

Additionally, we validated the user simulator via a small human assessment detailed in Appendix~\ref{app:human}.

\begin{table}[b]
\centering
\resizebox{0.5\linewidth}{!}{%
\setlength{\tabcolsep}{6pt}
\renewcommand{\arraystretch}{1.0}
\begin{tabular}{lccc}
\toprule
\textbf{Category} & \textbf{\#persona} & \makecell[b]{\textbf{\# validated} \\ \textbf{pairs}} & \makecell[b]{\textbf{Avg. required turns} \\ \textbf{ per template}} \\
\midrule
Informational & 30 & 215 & 6.60 \\
Instructional & 30 & 244 & 6.97 \\
Interactive   & 30 & 273 & 7.23 \\
Operational  & 30 & 221 & 6.48 \\
\midrule
\textbf{Total} & \textbf{30} & \textbf{953} & \textbf{6.85} \\
\bottomrule
\end{tabular}%
}
\caption{
Summary statistics of the evaluation dataset across four scenario categories. 
}
\label{tab:conversation_data}
\end{table}

\paragraph{Test models.} We evaluated a range of frontier LLMs from different providers: 
Gemini-2.5-pro, Gemini-2.5-flash, and Gemini-1.5-flash \citep{comanici2025gemini}; Claude Sonnet-4.5, Claude Haiku-4.5 and  Claude Sonnet-4 \citep{anthropic2024,anthropic2024claude3};  Qwen3-Next \citep{yang2025qwen3}, 
DeepSeek-R1-671B \citep{guo2025deepseek}.

\begin{table*}[t!]
\centering
\resizebox{0.85\linewidth}{!}{%
\begin{tabular}{lccccccccc}
\toprule
\multirow{2}{*}{\textbf{Test models}} & \multicolumn{4}{c}{\textbf{Task Completeness}} & \multicolumn{4}{c}{\textbf{User-Centric Personalization}} & \multirow{2}{*}{\textbf{Model}} \\
\cmidrule(lr){2-5} \cmidrule(lr){6-9}
 & ODI & DCA & FTP & \textbf{avg.} & AFM & OSF & SSA & \textbf{avg.} & \textbf{rating}\\
\midrule
Claude Sonnet 4 & 4.65 & 4.76 & 4.25 & 4.56 & 4.75 & 4.74 & 4.73 & 4.74 & 4.65 \\
Claude Sonnet 4.5 & 4.37 & 4.69 & 3.53 & 4.20 & 4.13 & 4.68 & 4.57 & 4.46 & 4.33 \\
Claude Haiku 4.5 & 4.02 & 4.64 & 3.16 & 3.94 & 4.47 & 4.57 & 4.61 & 4.55 & 4.25 \\
Qwen3-Next & 4.73 & 4.03 & 4.50 & 4.42 & 4.62 & 4.72 & 4.70 & 4.68 & 4.55 \\
DeepSeek-R1 & 4.68 & 4.65 & 4.59 & 4.64 & 4.60 & 4.54 & 4.54 & 4.56 & 4.60 \\
Gemini 2.5 Pro & \textbf{4.86} & \textbf{4.86} & \textbf{4.70} & \textbf{4.81} & \textbf{4.76} & \textbf{4.86} & \textbf{4.79} & \textbf{4.80} & \textbf{4.81} \\

Gemini 2.5 Flash & 4.79 & 4.72 & 4.12 & 4.60 & 4.70 & 4.72 & 4.76 & 4.75 & 4.68 \\

\bottomrule
\end{tabular}%
} 

\caption{Model performance across all rubrics at iteration $t=3$. (ODI: Output Delivery Integrity, DCA: Domain Conceptual Alignment, FTP: Functional Task Progression; AFM: Anticipatory Flow Management, OSF: Output Structure Fit, SSA: Sustained Style Adherence). See Appendix~\ref{app:additional_results} for results at $t=1,2$.}

\label{tab:model_metrics}
\vspace{-6pt}
\end{table*}

\paragraph{Rubrics design.}
We evaluate model performance using $K=6$ rubrics organized under two high-level dimensions: 
\emph{Task Completeness (TC)}, which measures whether the model understands the task intent and makes coherent progress toward the scenario objective, and 
\emph{User-Centric Personalization (UCP)}, which measures whether the model adapts to the user's preferences, identity, and interaction style throughout the dialogue. 
Each rubric is rated on a 1--5 scale, where 5 denotes the ideal outcome. 
While the rubric names are fixed, their descriptions and rating guidelines are iteratively refined through the co-evolution loop over $T=3$ iterations. 
The complete rubric definitions, including the six sub-rubrics and their rating criteria, are provided in Appendix~\ref{app:metrics}; implementation details are provided in Appendix~\ref{app:implementation}.

\paragraph{Evaluator-human alignment validation.} We conducted a human validation study to assess whether co-evolution improves evaluator--human alignment.
We sampled 40 persona-scenario pairs (10 per scenario category), had three annotators rate conversations across three co-evolution rounds using five-point scoring criteria and round-specific rubrics. 
Both human annotators and the LLM-as-a-judge applied identical rubrics in each round.
Therefore, this design enables us to test whether iterative refinement strengthens human-LLM alignment by directly comparing their respective scores for each conversation-rubric pair. Recruitment and additional study details are provided in Appendix~\ref{app:human}.

\subsection{Experiment Results and Analysis}
\subsubsection{Model Performance and Stratification}
Table~\ref{tab:model_metrics} details rating results across the two primary dimensions  at the end of the third iteration: (i) a model's ability to carry out a task over multiple turns, and (ii) its capacity to adapt responses to user-specific signals  while ensuring structural and stylistic consistency. 
The results reveal a clear stratification of model capabilities.

\textit{Gemini~2.5~Pro} leads the frontier with the highest model rating of 4.81, dominating all evaluation rubrics. 
This performance reflects that the model’s superior reasoning and planning capabilities translate into highly effective dialogue management over complex, multi-turn interactions.

By contrast, the other models show mixed performance, excelling in specific areas while lagging in others.
This variance underscores that high-quality multi-turn interaction requires more than isolated strengths; it demands a balanced integration of understanding, task progression, and user adaptation.

\paragraph{Task Completeness: understanding versus execution.}
On task-oriented metrics, \textit{Gemini~2.5~Pro} achieves the strongest overall Task Completeness performance, with an average score of $4.81$. 
This corresponds to a relative improvement of approximately $3.7\%$ over the best non-Gemini baseline (\textit{DeepSeek-R1}, $4.64$), and more than $22\%$ compared to smaller-capacity variants such as \textit{Claude Haiku~4.5}.
Crucially, this advantage is not confined to surface correctness: \textit{Gemini~2.5~Pro} also leads in Functional Task Progression (FTP), indicating a consistent ability to decompose tasks and track intermediate states across long-horizon interactions.

By contrast, runner-up models exhibit clear trade-offs between conceptual understanding and execution. 
For instance, \textit{the Claude Sonnet series} and \textit{DeepSeek-R1} achieve strong Domain Conceptual Alignment (DCA), yet their FTP scores lag behind \textit{Gemini~2.5~Pro} by $25\%$ or more in relative terms (e.g., \textit{Claude Sonnet~4.5}). 
Conversely, \textit{Qwen3-Next} prioritizes execution, with an FTP score within $4\%$ of \textit{Gemini~2.5~Pro}, but at the cost of substantially weaker DCA (approximately $17\%$ lower), suggesting brittle execution under conceptual ambiguity.
This execution gap is further magnified in reduced-capacity variants: \textit{Gemini~2.5~Flash} and \textit{Claude Haiku~4.5} trail \textit{Gemini~2.5~Pro} in FTP by roughly $12\%$ and $33\%$ respectively, highlighting the difficulty of sustaining long-horizon task progression without sufficient model capacity.

\paragraph{User-Centric Personalization: style consistency versus adaptability.}
\textit{Gemini~2.5~Pro} also leads in User-Centric Personalization, achieving an average score of $4.80$. 
Although the absolute margin over other frontier models is narrower in this domain, it still represents a consistent relative improvement of approximately $5\%$ over the strongest alternatives.
Its advantage is most pronounced in Anticipatory Flow Management (AFM), where it outperforms \textit{Claude Sonnet~4.5} by over $15\%$ and \textit{Claude Haiku~4.5} by roughly $6\%$, indicating a superior ability to anticipate evolving user intent rather than merely reacting to explicit cues.
Other frontier-scale models, including \textit{Claude Sonnet~4}, \textit{DeepSeek-R1}, and \textit{Qwen3-Next}, cluster closely in overall personalization performance, with relative differences largely within $5\%$.

In contrast, \textit{Claude Sonnet~4.5} demonstrates strong stylistic consistency, with OSF and SSA scores within $5\%$ of the best-performing model, but its AFM score lags by more than $13\%$, revealing a clear trade-off between style adherence and adaptability.
A similar pattern appears in \textit{Claude Haiku~4.5}, where stylistic alignment remains competitive while anticipatory adaptation degrades more sharply.
These results suggest that while stylistic personalization is relatively robust across models, adaptive personalization—particularly the ability to steer conversations in response to latent or shifting user intent—exhibits a stronger dependence on model capacity.

\begin{figure}[t]
    \centering
    \includegraphics[width=0.6\linewidth]{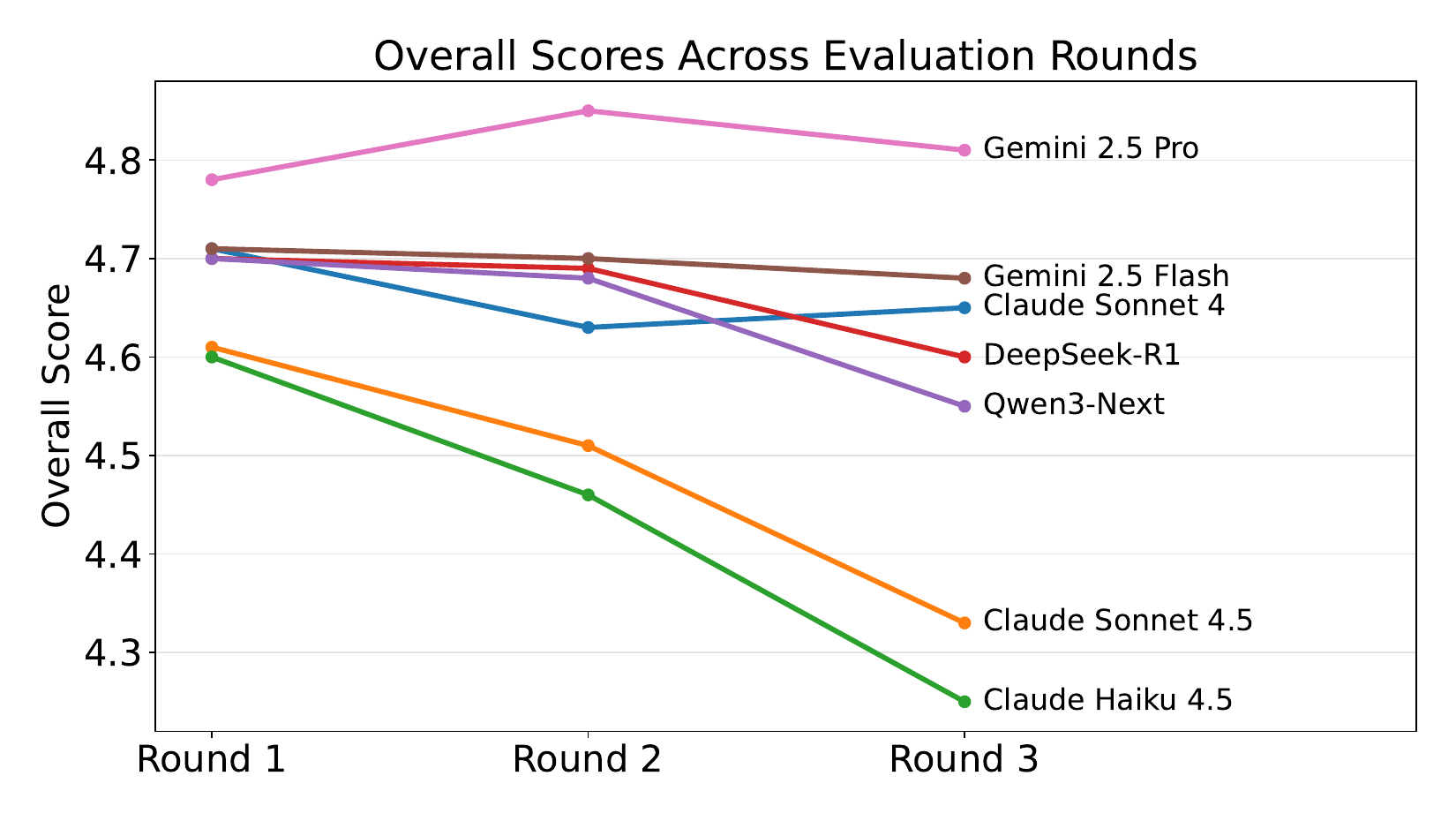}
    \caption{
    Model ratings across iterations. Iterative reflection improves model differentiation, yielding more distinct stratification and relative ordering.}
    \label{fig:reflection_slope}

\end{figure}

\begin{table}[t]
\centering
\small
\setlength{\tabcolsep}{5pt}
\renewcommand{\arraystretch}{1.2}
\begin{tabular}{lccc}
\toprule
\textbf{Metric} & \textbf{$t=1$} & \textbf{$t=2$} & \textbf{$t=3$} \\
\midrule
\multicolumn{4}{l}{\textit{Rubric Discriminability ($\Delta^{(t)} $)}} \\
Inter-model std.\ dev.\ ($\sigma_{\text{inter}}$)
    & 0.062 & 0.128 & \textbf{0.194} \\
\midrule
\multicolumn{4}{l}{\textit{Rubric Stability ($\Gamma^{(t)} $)}} \\
Intra-model variance ($\sigma^{2}_{\text{intra}}$)
    & 0.145 & 0.141 & \textbf{0.138} \\
Rank consistency (Spearman $\rho$)
    & 0.68 & 0.79 & \textbf{0.92} \\
\bottomrule
\end{tabular}
\caption{
Evolution of rubric discriminability and stability across iterations.  Rank consistency is the average Spearman correlation of model rankings across evaluation subsets within the same iteration. 
Iterative refinement increases inter-model separation while preserving
or improving scoring stability, validating the quantitative feedback
criteria used to accept rubric updates.
}
\label{tab:rubric_stats}
    \vspace{-6pt}        
\end{table}

\begin{table*}[t!]
\centering
\small
\setlength{\tabcolsep}{4pt}
\resizebox{0.85\linewidth}{!}{%
\begin{tabular}{llccccccccc}
\toprule
\multirow{2}{*}{\textbf{Round}} 
& \multirow{2}{*}{\textbf{Metric}}
& \multicolumn{4}{c}{\textbf{Task Completeness}} 
& \multicolumn{4}{c}{\textbf{User-Centric Personalization}} 
& \multirow{2}{*}{\textbf{Overall}} \\
\cmidrule(lr){3-6} \cmidrule(lr){7-10}
& & ODI & DCA & FTP & \textbf{avg.}
& AFM & OSF & SSA & \textbf{avg.}
& \\
\midrule
\multirow{2}{*}{1st-round}
& MAE $\downarrow$ 
& 0.68 & 0.54 & 0.76 & 0.66
& 0.71 & 0.83 & 0.69 & 0.74
& 0.70 \\
& QWK $\uparrow$ 
& 0.46 & 0.58 & 0.39 & 0.48
& 0.42 & 0.34 & 0.45 & 0.40
& 0.44 \\
\midrule
\multirow{2}{*}{2nd-round}
& MAE $\downarrow$ 
& 0.53 & 0.49 & 0.58 & 0.53
& 0.56 & 0.61 & 0.55 & 0.57
& 0.55 \\
& QWK $\uparrow$ 
& 0.58 & 0.62 & 0.51 & 0.57
& 0.54 & 0.52 & 0.56 & 0.54
& 0.56 \\
\midrule
\multirow{2}{*}{3rd-round}
& MAE $\downarrow$ 
& \textbf{0.48} & \textbf{0.47} & \textbf{0.52} & \textbf{0.49}
& \textbf{0.51} & \textbf{0.55} & \textbf{0.52} & \textbf{0.53}
& \textbf{0.51} \\
& QWK $\uparrow$ 
& \textbf{0.62} & \textbf{0.63} & \textbf{0.56} & \textbf{0.60}
& \textbf{0.58} & \textbf{0.57} & \textbf{0.58} & \textbf{0.58}
& \textbf{0.59} \\
\bottomrule
\end{tabular}%
}
\caption{
Human validation of LLM-as-judge alignment across co-evolution rounds.
MAE measures absolute discrepancy against averaged human ratings, while QWK measures ordinal agreement with individual human annotators on the 1--5 rubric scale and is then averaged across annotators.
Lower MAE and higher QWK indicate stronger alignment with human judgments.
}
\label{tab:human_alignment_full}
\vspace{-10pt}
\end{table*}

\subsubsection{Effect of Co-Evolution Iterations}
\label{sec:effect_iterations}

Figure~\ref{fig:reflection_slope} shows that model ratings are highly clustered in the initial round, limiting reliable model ordering. 
After iterative reflective rubric refinement and co-evolutionary user simulation, later iterations reveal clearer and more fine-grained stratification of model performance, suggesting that CoReflect progressively uncovers subtle quality differences that are initially obscured. 
This trend is quantitatively supported by Table~\ref{tab:rubric_stats}: rubric discriminability $\Delta^{(t)}$, measured by inter-model rating dispersion, increases monotonically across iterations, improving the ability to distinguish models with otherwise similar ratings, while rubric stability $\Gamma^{(t)}$ is maintained or slightly improved, as reflected by stable intra-model variance and improved rank consistency. 
These results indicate that CoReflect strengthens evaluative signal without introducing additional stochastic noise.

Our human validation study assesses whether co-evolution improves evaluator-human alignment.
We report MAE for absolute score discrepancy against averaged human ratings, and quadratic weighted kappa (QWK) for ordinal agreement on the 1--5 rubric scale.
For QWK, we compute agreement with each individual annotator and then average the scores, avoiding additional discretization of averaged human ratings.
As shown in Table~\ref{tab:human_alignment_full}, evaluator-human alignment improves across co-evolution rounds, with the largest gain observed from the 1st to the 2nd round and a smaller improvement from the 2nd to the 3rd round. 
This trend suggests that early reflection substantially corrects coarse evaluator misalignment, while later iterations provide more incremental refinements. 
The improvement is not uniform across rubrics: structurally sensitive dimensions such as OSF and task-progression dimensions such as FTP show larger reductions in MAE, whereas DCA starts with relatively stronger agreement and therefore leaves less room for improvement. 
Overall, MAE decreases from 0.70 to 0.51 and QWK increases from 0.44 to 0.59, indicating that CoReflect improves both score calibration and ordinal agreement with human judgments.

\begin{figure}[t]
    \centering
    \includegraphics[width=0.8\linewidth]{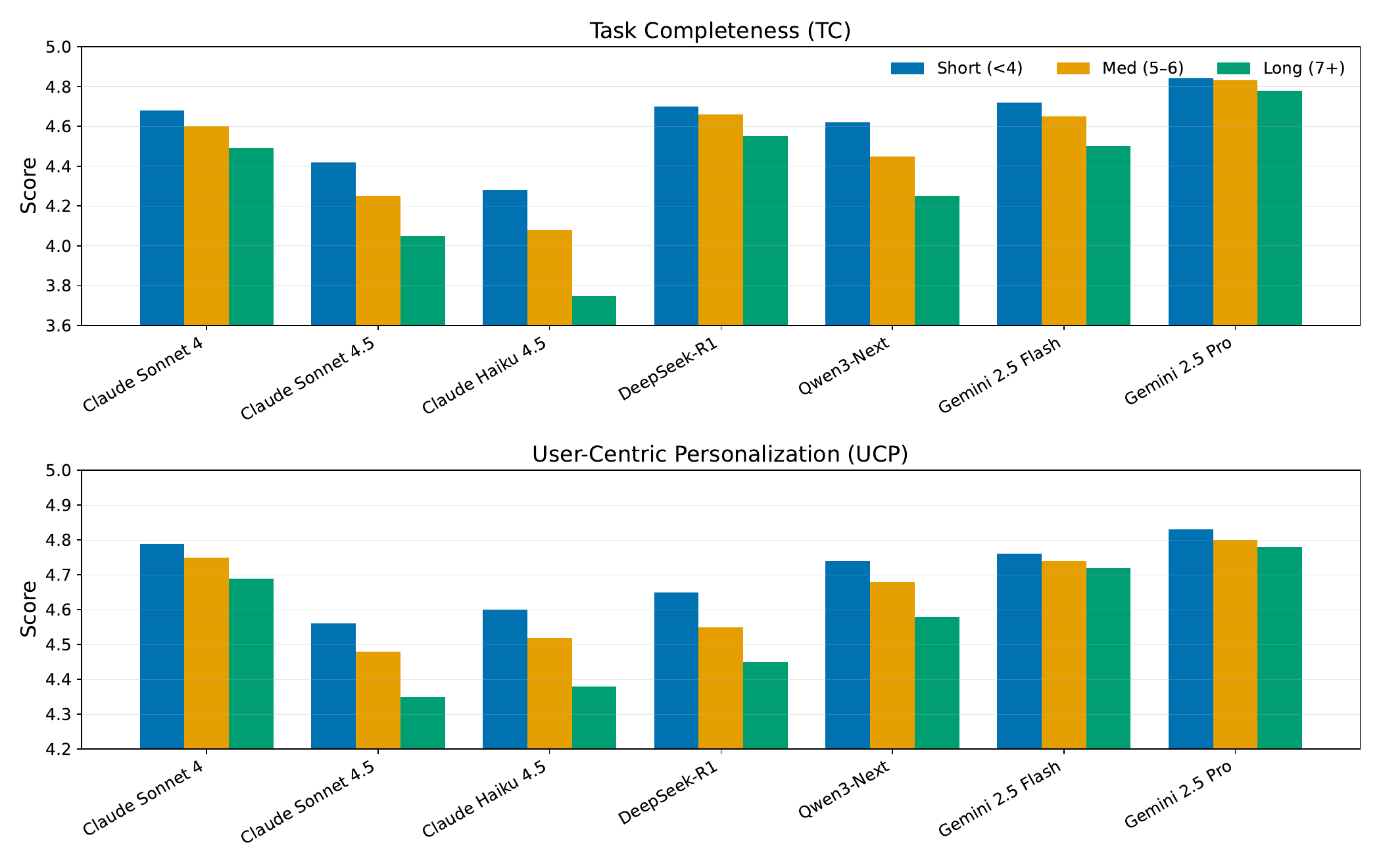}
    \caption{
    Model ratings across Task Completeness and User-Centric Personalization, stratified by conversation length.
    }
    \label{fig:length_impact}

\end{figure}

\subsubsection{Effect of Conversation Length}
Figure~\ref{fig:length_impact} illustrates the impact of conversation length on model performance. Across all models, both Task Completeness and User-Centric Personalization scores decline as conversation length increases, indicating that extended interactions pose a systematic challenge for current models.

Task completeness degrades more significantly than personalization in long-horizon dialogues.
While the models maintain stable personalization, their ability to track task states falters over time. 
This trend widens the performance gap between large-scale and lightweight models, which is indistinguishable in shorter interactions.

These results suggest conversation length as a critical factor for long-horizon reasoning and execution, highlighting the necessity of length-aware analysis in multi-turn conversational benchmarks to accurately differentiate model capabilities.

\subsubsection{Case Study}

\begin{figure*}
    \centering

    \includegraphics[width=0.8\linewidth]{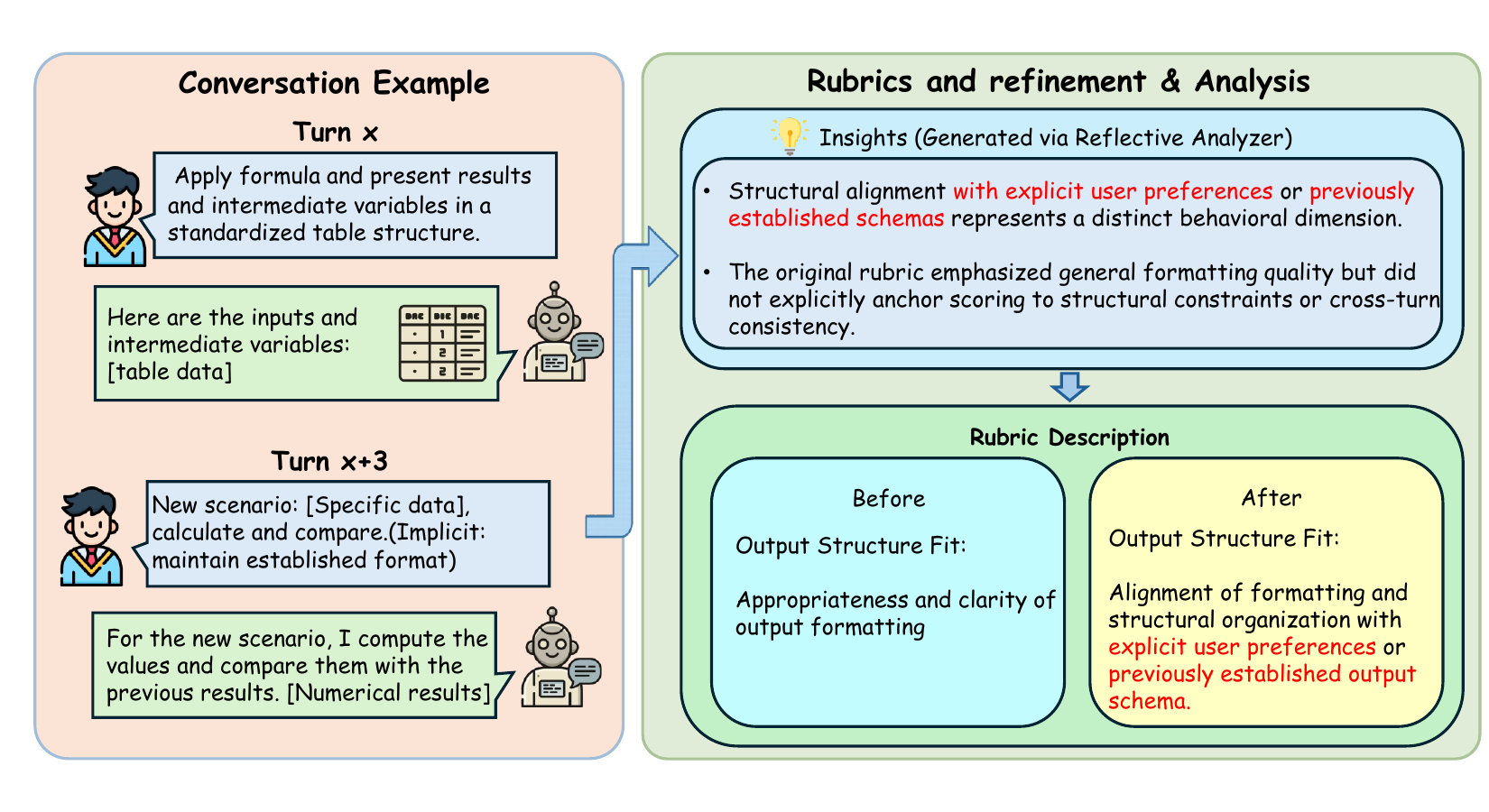}

\caption{
Example of reflective rubric refinement for identifying cross-turn structural regressions. The model initially follows a user-requested table schema but later reverts to inline narrative text while remaining semantically correct. Based on recurring observations of this pattern, the reflective analyzer refines the \textit{Output Structure Fit} rubric to explicitly evaluate alignment with user preferences and previously established output structures. The associated five-point scoring criteria are also refined accordingly.
}
    \label{fig:case_study}
 
\end{figure*}

Figure~\ref{fig:case_study} illustrates a representative example of reflective rubric refinement. In the conversation, the user first requests the model to present intermediate variables and results in a standardized table format. While the model initially follows the required structure, it later reverts to inline narrative text in a subsequent turn despite remaining semantically correct. Similar structural regressions were repeatedly observed across diverse multi-turn scenarios.

Based on these recurring patterns, the reflective analyzer identifies cross-turn structural consistency as an important behavioral dimension not fully captured by the original rubric. Consequently, the framework refines the \textit{Output Structure Fit} rubric from emphasizing general formatting quality to explicitly evaluating alignment with user preferences and previously established output schemas. In addition to updating the rubric description, the framework also refines the associated five-point scoring criteria to better distinguish varying levels of structural consistency, although the detailed scale changes are omitted here for brevity.

This example demonstrates how the proposed framework progressively improves evaluation granularity by incorporating recurring behavioral insights discovered during multi-turn interactions.
\section{Related Work}
    \vspace{-5pt}
\subsection{Structured User Simulation}
    \vspace{-5pt}
Structured simulations provide a controllable framework for systematically probing model behavior, ranging from classical planning-based systems~\citep{reiter2000building} to modern neural plan-and-write pipelines~\citep{yao2019plan,xu2020megatron}. By applying these techniques to conversational evaluation, persona-driven and template-based simulations enable targeted stress-testing of a model’s personalization, consistency, and task execution.


\subsection{LLM-as-a-Judge and Scalable Evaluation}
To scale evaluation beyond human annotation, LLM-as-a-judge has emerged as a widely adopted paradigm. G-Eval~\citep{liu2023geval} demonstrates that LLM-based evaluators can achieve strong correlation with human judgments, while MT-Bench~\citep{zheng2023mtbench} and \textsc{Chatbot Arena}~\citep{chiang2024chatbotarena} enable large-scale multi-turn comparison through pairwise or rubric-based scoring. Complementary benchmarks such as LongBench~\citep{bai2024longbench}, L-Eval~\citep{an2024eval}, and RULER~\citep{hsieh2024ruler} focus on long-context understanding and sustained reasoning. Despite their effectiveness, most LLM-based evaluation frameworks adopt static rubrics or prompts, making them less sensitive to newly emerging failure modes in personalized and long-horizon interactions.

\subsection{Self-Refining Evaluation}

Motivated by the limitations of static LLM-as-a-judge frameworks, recent work has begun to explore \emph{self-refining} evaluation paradigms that improve evaluators through reflection and iterative feedback loops. A representative line of research studies reflective evaluation, where LLM-based judges generate structured rationales or critiques before assigning scores, helping surface latent failure modes beyond scalar judgments \citep{liu2023geval,zheng2023mtbench,shinn2023reflexion}. Building on such signals, other work introduces iterative refinement loops that update evaluation prompts, rubrics, or protocols based on low-performing or ambiguous cases. In conversational settings, LLM-based user simulators enable interactive evaluation beyond fixed contexts: \citet{wang2023rethinking} employ a ChatGPT-based simulator for multi-turn conversational recommendation evaluation, while \citet{liu2023must} show that using multiple simulators improves robustness over relying on a single one. Relatedly, \citet{hu2022unlocking} use LLM-simulated user feedback as an annotation-free signal to iteratively optimize dialogue responses. Together, these studies highlight a shift toward closed-loop evaluation frameworks that continuously refine evaluation criteria as model behaviors evolve.

    \vspace{-5pt}
\subsection{Evaluation of Personalized Dialogue} Personalization in conversational AI has traditionally centered on persona grounding, preference modeling, and adaptive generation. Foundational benchmarks like \textsc{Persona-Chat}~\citep{zhang2018personalizing}, \textsc{BlendedSkillTalk}~\citep{smith2020blend}, and \textsc{EmpatheticDialogues}~\citep{rashkin2019empathetic} established controlled environments for assessing consistency, multi-skill agility, and emotional grounding. Recent efforts, such as  DialogBench~\citep{ou2024dialogbench} and PersonaConvBench~\citep{li2025personalized}, have extended these evaluations to multi-turn interactions. However, while these resources effectively identify personalization failures, they rely on static scripts and evaluation criteria that struggle to adapt as conversational behaviors and user expectations evolve.

\section{Conclusion}

We introduced CoReflect, an autonomous framework unifying persona-driven simulation with iterative rubric refinement. 
By dynamically evolving evaluation criteria based on observed behaviors, CoReflect captured nuances that static benchmarks tend to miss.
Experiments confirmed this self-refining approach significantly improved discriminatory resolution, resulting in clear, high-fidelity model stratification.

CoReflect provides a scalable solution for complex, hard-to-define domains such as personalization by automatically updating obsolete criteria and mitigating the subjective bias inherent in human-designed evaluation architectures. Future work will aim to strengthen CoReflect by grounding simulation in real user interactions and improving evaluation robustness through richer, bias-aware judging and rubric design.

\bibliography{main}

\appendix

\section*{Appendix}

\section{ User Persona Definition}
\label{app:a}
To simulate diverse user profiles in personalized multi-turn conversations, we define a structured persona schema capturing both the user’s expressive behavior and their expectations for AI responses. Each persona is composed of the following elements:

\paragraph{ID and Role} A unique identifier and a brief description of the user’s background or profession (e.g., University Student, Senior Developer). This provides contextual grounding for the user’s domain knowledge and communication style.

\paragraph{Language} The primary language used by the persona, enabling multilingual dialogue simulation and evaluation.

\paragraph{Personality} This component models how the user naturally communicates. It includes:
\begin{itemize}
    \item \textbf{Traits:} Core personality characteristics that influence conversational behavior (e.g., Impatient, Inquisitive, Skeptical). 
    \item \textbf{Tone:} The persona’s typical emotional tone in conversation (e.g., Formal, Sarcastic, Friendly).
    \item \textbf{Verbosity:} Typical message length and level of detail (e.g., Terse, Verbose).
    \item \textbf{Quirks:} Specific linguistic habits that define the user’s style (e.g., Skips pleasantries, Uses all lowercase, Frequently references personal projects).
\end{itemize}

\paragraph{Preferred Response Style} This defines how the user expects the AI assistant to respond and serves as the personalization target. It includes preferences for:
\begin{itemize}
    \item Tone (e.g., Empathetic, Direct),
    \item Verbosity (e.g., Concise, Detailed),
    \item Reasoning Depth (e.g., Brief rationale, Step-by-step explanation),	
    \item Engagement (e.g., Neutral, Asks follow-ups),
    \item Clarity (e.g., Simple and clear, Technically precise).
\end{itemize}

\section{Scenario Category Definition}
\label{app:b}
To comprehensively evaluate the capabilities of large language models in multi-turn dialogue, we define four broad categories of interaction scenarios and a structured schema for their definition.

\paragraph{Scenario Categories}
\begin{itemize}
    \item \textbf{Instructional Dialogue:} Encompasses interactions where the user’s objective is to acquire new knowledge or develop specific skills. The model serves in an instructional capacity, providing pedagogical guidance and step-by-step reasoning.
    \item \textbf{Informational Dialogue:} Characterized by the user’s intent to obtain accurate, context-relevant information. The model functions as a factual resource, delivering concise and verifiable responses.
    \item \textbf{Operational Dialogue:} Includes interactions where the user seeks assistance in completing a concrete task or producing a functional output. The model is employed as a productivity facilitator to generate content or perform structured actions.
    \item \textbf{Interactive Dialogue:} Captures scenarios where the user’s primary aim is social, emotional, or imaginative engagement. The model adopts a human-like persona, prioritizing coherence, tone adaptation, and creative collaboration.
\end{itemize}

\paragraph{Scenario Definition Schema}
\begin{itemize}
    \item \textbf{ID and Title:} A unique scenario identifier and a concise title summarizing the context (e.g., Planning a Weekly Study Schedule, Summarizing Technical Emails).
    \item \textbf{Category:} A high-level classification of the task domain, selected from predefined types such as Work \& Productivity, Learning \& Study, Creative \& Brainstorming, Personal \& Planning, and Information \& Factual.
    \item \textbf{Situation Description:} A neutral, third-person description of the background or external context.
    \item \textbf{Core Task:} A focused objective to be accomplished (e.g., Organize a calendar with overlapping deadlines).
    \item \textbf{Expected Interaction:}
    \begin{itemize}
        \item Turn Complexity: Short (<4 turns), Medium (5–6 turns), or Long (7+ turns).
        \item Dialogue Flow Type: Goal-Oriented, Open-Ended, Step-by-Step Guidance, or Collaborative Generation.
    \end{itemize}
    \item \textbf{Success Criteria:} A neutral description of what constitutes a successful outcome (e.g., a clear and logically ordered prioritization of incoming requests).
\end{itemize}

\section{ Automated Consistency Check}
\label{app:c}
A critical validation step is performed before any persona is paired with a scenario. To prevent the creation of illogical or nonsensical conversations, we employ an automated consistency check using a large language model (LLM) configured as a ``judge.''  

\paragraph{Protocol Steps}
\begin{enumerate}
    \item \textbf{Framing the Prompt:} The function sends a targeted prompt to an LLM, instructing it to act as a ``logical reasoning expert'' for the specific task of evaluating a pairing's plausibility.
    \item \textbf{Providing Context:} The prompt is populated with the role from the persona definition and the core\_task and title from the scenario definition.
    \item \textbf{Binary Decision:} The LLM is strictly constrained to answer with only a single word, ``Yes'' or ``No.''
    \item \textbf{Filtering:} If the LLM returns ``Yes,'' the pair is approved. If ``No,'' the combination is logged and discarded.
\end{enumerate}

\section{Conversation Simulation and Template}
\label{app:d}
To simulate realistic multi-turn conversations between a user and an AI assistant, we define a structured generation protocol. Each conversation is generated from a persona-scenario pair and is designed to assess both personalization alignment and task adherence.

\paragraph{Protocol Components}
\begin{itemize}
    \item \textbf{Persona Context:} Full instantiation of tone, verbosity, quirks, and preferences.
    \item \textbf{Scenario Context:} Neutral task description from scenario schema.
    \item \textbf{Generation Instructions:}
    \begin{itemize}
        \item \texttt{turnType} (Early Turn, Intermediate Turn, Challenging Turn),
        \item \texttt{turnIntent} (logic-level description of user’s aim),
        \item \texttt{instructionForEval} (rationale for why this turn tests AI behavior).
    \end{itemize}
    \item \textbf{Conversation Structure:}
    \begin{itemize}
        \item Early Turn: Persona initiates task using typical style.
        \item Normal Turns: Task progression through elaboration or clarification.
    \end{itemize}
\end{itemize}

\section{Metrics Design}
\label{app:metrics}

We evaluate model performance along two high-level dimensions: 
(i) \emph{Task Completeness (TC)}, which captures whether a model correctly understands task intent and makes coherent progress toward fulfilling the scenario objective; and 
(ii) \emph{User-Centric Personalization (UCP)}, which reflects the model's ability to adapt its responses to a user's identity, preferences, and interaction style over the course of a dialogue.

Each dimension is assessed using three complementary rubrics, yielding a total of $K=6$ rubrics:

\begin{itemize}[leftmargin=*]
     \item \textbf{Task Completeness}: 
     Domain Conceptual Alignment (DCA), Output Delivery Integrity (ODI), and Functional Task Progression (FTP) evaluate whether responses are conceptually correct, actionable, and incrementally advance the task.

     \item \textbf{User-Centric Personalization}: 
     Anticipatory Flow Management (AFM), Output Structure Fit (OSF), and Sustained Style Adherence (SSA) assess whether the model anticipates user needs, adapts response organization, and maintains consistent stylistic alignment.
\end{itemize}

Each rubric is rated on a 1--5 scale, where higher scores indicate better performance. 
The rubric names remain fixed across iterations to ensure comparability, while their descriptions and rating guidelines are refined through the co-evolution loop. 
In our experiments, we conduct $T=3$ refinement iterations.

\paragraph{Meta-Metrics}
\begin{itemize}[noitemsep, topsep=0pt, leftmargin=*]
    \item \textbf{User-Centric Personalization:} Effectiveness of adapting to the user’s persona, dialogue history, and explicit/implicit feedback.
    \item \textbf{Task \& Completeness:} Degree to which the model fulfills the task and adheres to all stated requirements/constraints.
\end{itemize}

\paragraph{Personalization Sub-metrics (Refined via Self-Reflection)}
\begin{itemize}[noitemsep, topsep=0pt, leftmargin=*]
    \item \textbf{Sustained Style Adherence:} Consistency in maintaining adapted style and honoring negative constraints across turns.
    \item \textbf{Output Structure Fit:} Alignment of formatting/structure with user preferences or requested schema.
    \item \textbf{Anticipatory Flow Management:} Proactive pacing and next-step guidance that matches user needs and context.
\end{itemize}

\paragraph{Task Execution Sub-metrics (Refined via Self-Reflection)}
\begin{itemize}[noitemsep, topsep=0pt, leftmargin=*]
    \item \textbf{Domain Conceptual Alignment:} Correct identification and sustained adherence to the task’s conceptual domain (e.g., fictional vs.\ real, technical vs.\ general).
    \item \textbf{Output Delivery Integrity:} Reliable, complete delivery of required artifacts without truncation/corruption (how it delivers, not what).
    \item \textbf{Functional Task Progression:} Multi-step advancement from understanding to precise, actionable outputs.
\end{itemize}

Here we provide the detailed 5-point scoring rubrics.
\paragraph{Meta-Metrics}
\textbf{User-Centric Personalization (Score: 1–5)}  
\begin{itemize}
    \item 5 (Exceptional): Flawless personalization—highly adaptive and anticipatory with no missed cues or lapses.
    \item 4 (Strong): Good personalization but with one noticeable miss (e.g., slight verbosity, delayed adaptation).
    \item 3 (Moderate): Some effort at personalization, but with inconsistencies or over-reliance on user correction.
    \item 2 (Weak): Minimal personalization; the model sounds generic, forgets earlier info, or needs multiple corrections.
    \item 1 (Poor): No sign of personalization—repetitive, forgetful, or misaligned responses.
\end{itemize}

\textbf{Task \& Completeness (Score: 1–5)}  
\begin{itemize}
    \item 5 (Exceptional): Perfect task completion, honoring all constraints, with no filler or irrelevant content.
    \item 4 (Strong): Task met, but with one issue—e.g., slight verbosity, missed minor constraint.
    \item 3 (Moderate): Goal reached but required user correction or included multiple avoidable errors.
    \item 2 (Weak): Task only partially met—critical information missing or misunderstood.
    \item 1 (Poor): Task clearly failed or severely misinterpreted.
\end{itemize}

\paragraph{Personalization Sub-Metrics}
\textbf{Sustained Style Adherence (1–5)}  
\begin{itemize}
    \item 5: Flawlessly maintains the adapted style and all negative constraints throughout.
    \item 4: Maintains style well, with only one minor slip.
    \item 3: Shows effort but style is inconsistent or requires user correction.
    \item 2: Frequently deviates or repeatedly violates constraints.
    \item 1: No sustained adherence; style is erratic.
\end{itemize}

\textbf{Output Structure Fit (1–5)}  
\begin{itemize}
    \item 5: Perfectly adapts formatting and structure to the user's needs.
    \item 4: Good adaptation with one minor inconsistency.
    \item 3: Some adaptation, but formatting is inconsistent or generic.
    \item 2: Little to no adaptation; ignores cues.
    \item 1: No adaptation; formatting misaligned.
\end{itemize}

\textbf{Anticipatory Flow Management (1–5)}  
\begin{itemize}
    \item 5: Proactively anticipates user’s next steps, natural and efficient flow.
    \item 4: Generally anticipates needs well, with one slightly delayed transition.
    \item 3: Attempts to manage flow but sometimes misjudges.
    \item 2: Rarely anticipates; conversation feels disjointed.
    \item 1: Disrupts flow by consistently misinterpreting user intent.
\end{itemize}

\paragraph{Task Execution Sub-Metrics}
\textbf{Domain Conceptual Alignment (1–5)}  
\begin{itemize}
    \item 5: Perfectly identifies and operates within domain.
    \item 4: Good alignment, one minor self-corrected drift.
    \item 3: Some alignment but with inconsistencies.
    \item 2: Frequent misalignments or wrong domain.
    \item 1: Complete failure to align with domain.
\end{itemize}

\textbf{Output Delivery Integrity (1–5)}  
\begin{itemize}
    \item 5: Flawless and complete delivery with no truncation.
    \item 4: Full output with one minor glitch.
    \item 3: Mostly complete but some truncation or errors.
    \item 2: Significant truncation or multiple errors.
    \item 1: Severely truncated or failed delivery.
\end{itemize}

\begin{table*}[t!]
\centering
\scalebox{0.72}{
\begin{tabular}{p{0.5in}p{7.5in}}
\toprule
\multicolumn{2}{p{5in}}{
\textbf{Example of template prompt with persona and scenario context}:}\\
\midrule
\textbf{Context block}
 & 
\textbf{ConversationID:} P001\_S01\newline
\textbf{Persona (P001):} Senior developer; skeptical, impatient, analytical; direct and terse communication; uses technical jargon and frequently challenges the AI's accuracy.\newline
\textbf{Scenario (S01):} Instructional dialogue explaining photosynthesis; user seeks a clear, step-by-step explanation of inputs, outputs, and the two main stages (light-dependent and light-independent reactions).\\

\midrule

\textbf{Template} &

\textbf{Turn 1 (Early Turn)}\newline
\textbf{TurnIntent:} Initiate Core Task\newline
\textbf{Instruction:} Evaluate the AI's ability to provide a structured, high-level overview and a specific chemical formula as requested. The response should be direct and technical.\newline

\textbf{Turn 2 (Normal Turn)}\newline
\textbf{TurnIntent:} Request Specific Detail\newline
\textbf{Instruction:} Assess the AI's ability to explain a sub-process involving ATP and NADPH with technical accuracy while remaining concise.\newline

\textbf{Turn 3 (Normal Turn)}\newline
\textbf{TurnIntent:} Challenge for Deeper Detail\newline
\textbf{Instruction:} Test the AI's understanding of RuBisCO and carbon fixation, ensuring precision in the biochemical description.\newline

\textbf{Turn 4 (Preference Recall Turn)}\newline
\textbf{TurnIntent:} Request Summary \& Recall Preference\newline
\textbf{Instruction:} Check whether the AI recalls the explicit preference for the “balanced chemical equation” stated in Turn 1 and can summarize the steps coherently.\newline

\textbf{Turn 5 (Normal Turn)}\newline
\textbf{TurnIntent:} Validate Accuracy with Edge Case\newline
\textbf{Instruction:} Evaluate whether the AI correctly handles a nuanced biological question involving plant physiology without oversimplification.\\

\midrule

\textbf{Usage} &
This template illustrates how persona traits and scenario context guide turn-level evaluation.  
Each turn specifies an intent and an instruction for evaluation, enabling fine-grained assessment of AI behavior in personalized multi-turn dialogue. \\
\bottomrule
\end{tabular}
}
\caption{Example template prompt used in personalized multi-turn conversation evaluation, incorporating essential persona and scenario context.}
\label{tab:example_template_prompt}
\end{table*}

\textbf{Functional Task Progression (1–5)}  
\begin{itemize}
    \item 5: Autonomously drives task, flawless synthesis into actionable output.
    \item 4: Good progression with one minor error.
    \item 3: Progresses task but needs nudging and has errors.
    \item 2: Requires constant prompting; output has major errors.
    \item 1: Fails to progress task.
\end{itemize}

\newpage
\section{Measures for Rubrics Refinement}
\label{app:rubric_metrics}

We formalize two measures to quantify the effect of rubric refinement on evaluation quality.

\paragraph{Rubric discriminability.}
At iteration $t$, rubric discriminability is defined as the standard deviation of models' mean ratings:
\[
\Delta^{(t)} \triangleq \sqrt{\frac{1}{M - 1} \sum_{j=1}^{M} \left(\mu_j^{(t)} - \bar{\mu}^{(t)}\right)^2},
\]
where $\mu_j^{(t)}$ denotes the mean rating of model $j$, and
\[
\bar{\mu}^{(t)} \triangleq \frac{1}{M} \sum_{j=1}^{M} \mu_j^{(t)}
\]
is the average rating across all $M$ models. Higher values of $\Delta^{(t)}$ indicate stronger separation between model capabilities.

\paragraph{Rubric stability.}
We assess rubric stability using two complementary metrics.

\textit{(i) Intra-model variance.}
We define intra-model variance as the average within-model variance of ratings across conversations:
\[
\Gamma^{(t)} \triangleq \frac{1}{M} \sum_{j=1}^{M} \left(\sigma_j^{(t)}\right)^2,
\]
where $\sigma_j^{(t)}$ denotes the standard deviation of ratings for model $j$. Lower values indicate more consistent evaluations.

\textit{(ii) Rank consistency.}
We compute the Spearman rank correlation coefficient $\rho$ to measure the consistency of model rankings across different evaluation subsets. Higher $\rho$ values indicate that the rubric preserves the relative performance ordering of models despite variations in conversation content.

\section{Implementation Details}
\label{app:implementation}

\paragraph{Model access and infrastructure.}
We conducted all experiments using a combination of Google AI Studio\footnote{https://aistudio.google.com/} and Vertex AI\footnote{https://cloud.google.com/vertex-ai}. 
Gemini models were accessed via Google AI Studio, while all non-Gemini models were accessed through Vertex AI.
To ensure broad coverage of contemporary model families and reasoning capabilities, we evaluated the following state-of-the-art models:
\begin{itemize}[leftmargin=*]
    \item \texttt{claude-sonnet-4@20250514}
    \item \texttt{claude-sonnet-4-5@20250929}
    \item \texttt{claude-haiku-4-5@20251001}
    \item \texttt{deepseek-ai/deepseek-r1-0528-maas}
    \item \texttt{gemini-2.5-flash@20250617}
    \item \texttt{gemini-2.5-pro@20250617}
\end{itemize}

\paragraph{Model roles within CoReflect.}
Within the CoReflect framework, different large language models were assigned fixed functional roles according to their capabilities.
\emph{Gemini~2.5~Flash} served as the backbone model for both the user simulator and the LLM-as-a-judge evaluator, where efficiency, stability, and responsiveness in multi-turn interaction and scoring were critical.
\emph{Gemini~2.5~Pro} was used as the reflective analyzer, which required stronger reasoning ability to synthesize evaluation outcomes, identify systematic behavioral patterns, and propose updates to both rubrics and conversation planning strategies.

\paragraph{Co-evolution procedure.}
CoReflect operates through an iterative co-evolution process in which dialogue simulation, evaluation, and rubric refinement are repeatedly coupled.
Across experiments, we performed three refinement iterations in total.
In each iteration, the conversation planner generated structured interaction templates that guided the user simulator, whose resulting conversations were evaluated by the LLM-as-a-judge under the current rubric definitions.
The reflective analyzer then aggregated evaluation signals across conversations and models to identify systematic weaknesses and emerging behaviors, which were used to update both rubric descriptions and planner templates before the next iteration.

\section{Evaluation Cost and Cost–Quality Trade-off}

Longer and more detailed rubrics may improve evaluation accuracy but also increase the cognitive cost for human evaluators and token cost for LLM judges. We therefore examine whether the additional evaluation information justifies this overhead. Table~\ref{tab:evaluation_cost} reports the average number of conversation and rubric tokens that human raters need to read in each iteration, together with human-alignment performance.

\begin{table}[!ht]
\centering
\small
\resizebox{0.6\linewidth}{!}{
\begin{tabular}{c|ccc|c|cc}
\hline
R & C. Tok. & R. Tok. & T. Tok. & $\Delta$T & MAE$\downarrow$ & QWK$\uparrow$ \\
\hline
1 & 5908 & 1376 & 7284 & -- & 0.70 & 0.44 \\
2 & 5559 & 1765 & 7324 & +0.55\% & 0.55 & 0.56 \\
3 & 5798 & 2191 & 7989 & +9.67\% & 0.51 & 0.59 \\
\hline
\end{tabular}
}
\caption{Evaluation cost and human-alignment performance across CoReflect iterations. C., R., and T. denote conversation, rubric, and total tokens, respectively.}
\label{tab:evaluation_cost}
\end{table}

Although the rubric becomes more detailed across iterations, the overall human-facing evaluation burden increases only moderately. From Round 1 to Round 3, the total number of tokens increases by 9.67\%, indicating that the additional rubric detail does not substantially inflate annotation-facing input.

For the full LLM-based evaluation pipeline, the marginal cost of rubric refinement is even smaller, since most computation comes from iterative multi-turn conversation generation across models and scenarios. The refined rubric only adds a bounded increase to the judge input. Importantly, this additional cost is accompanied by improved human alignment: MAE decreases from 0.70 to 0.51, while QWK increases from 0.44 to 0.59. Overall, these results suggest a favorable cost--quality trade-off: modest additional evaluation context yields substantially clearer and more reliable evaluation criteria.

\section{Human Evaluation Studies}
\label{app:human}

\begin{figure}[t]
    \centering
    \includegraphics[width=0.7\linewidth]{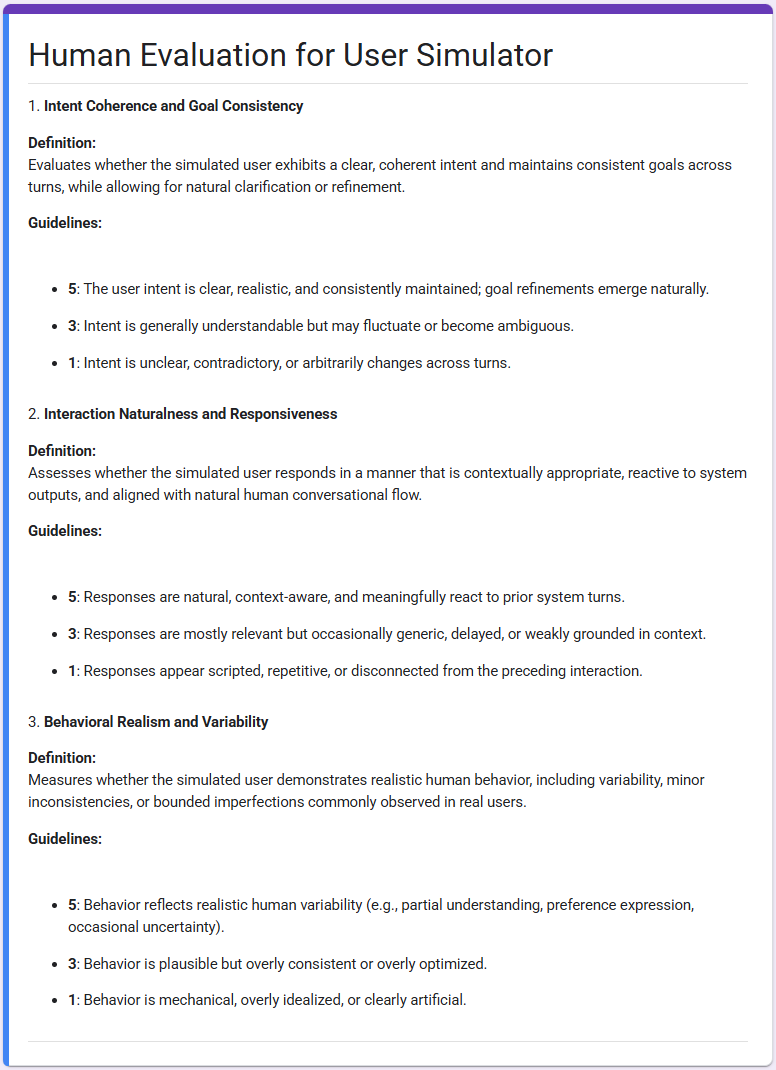}
    \caption{Human evaluation form used to assess the quality of simulated user behavior.}
    \label{fig:forms}
\end{figure}

We conducted two human evaluation studies with three graduate-student annotators recruited from a university in the United States. Annotators were compensated \$15 for the simulator validation tasks and \$30 for the conversation rating tasks. 
These amounts were determined based on the expected workload and were considered reasonable compensation for graduate-student participants in the region. Prior to participation, annotators were informed about the purpose of the study, the evaluation procedure, and how their annotations would be used for research, and consent was obtained before data collection. We  elucidate our evaluation design and results for the conversation rating tasks in Section~\ref{sec:exp}; here we detail our methodology and outcomes of the user simulator validation.

In the validation study on our user simulator, each annotator independently assessed ten randomly sampled conversations generated by the user simulator, resulting in a total of 30 ratings per evaluation aspect. Ratings were provided on a 5-point Likert scale along three dimensions: (1) \textit{Intent Coherence and Goal Consistency}, (2) \textit{Interaction Naturalness and Responsiveness}, and (3) \textit{Behavioral Realism and Variability}. The evaluation form used in this study is shown in Figure~\ref{fig:forms}.

Given the inherent subjectivity of human judgments, we measured inter-annotator agreement using Fleiss’ Kappa ($\kappa$). The average agreement score across all metrics was $\kappa = 0.68$, corresponding to \emph{substantial agreement}. This level of consistency indicates that the ratings reflect a shared assessment of simulator quality rather than random variation across annotators.

\begin{table}[t]
\centering
\small
\setlength{\tabcolsep}{3pt}  
\renewcommand{\arraystretch}{1.15}
\begin{tabular}{p{2.9cm}cc}
\toprule
\textbf{Evaluation Metric} & \textbf{Rating (Mean $\pm$ Std)} & \textbf{Kappa ($\kappa$)} \\
\midrule
Intent Coherence \& Consistency   & 4.53 $\pm$ 0.63 & 0.72 \\
Interaction Naturalness           & 4.27 $\pm$ 0.78 & 0.65 \\
Behavioral Realism \& Variability & 4.10 $\pm$ 0.84 & 0.61 \\
\bottomrule
\end{tabular}
\caption{
Human evaluation results ($N=30$ ratings per metric). Scores are reported on a 5-point Likert scale. Fleiss’ Kappa values above 0.60 indicate substantial inter-annotator agreement.
}
\label{tab:user_human_eval}
\end{table}

Quantitative results are summarized in Table~\ref{tab:user_human_eval}. Across all three dimensions, the simulator achieved consistently high mean scores, with strong agreement among annotators, supporting its effectiveness in producing realistic and human-like user behavior.

\section{Use of AI Assistants}
In accordance with the ACL Publication Ethics Policy, we did not employ AI assistants to generate the initial draft or the core innovative ideas of this paper. 
We utilized AI tools to perform editing for improved fluency and grammatical correctness. 
Beyond these applications and the specific experimental uses detailed in the methodology, such as synthetic data generation and model testing, no other AI tools were used in the research process.

\section{Additional Experimental Results}
\label{app:additional_results}
\begin{table*}[h!]
\centering
\resizebox{0.9\linewidth}{!}{%
\begin{tabular}{lccccccccc}
\toprule
\multirow{2}{*}{\textbf{Test Models}} & \multicolumn{4}{c}{\textbf{Task Completeness}} & \multicolumn{4}{c}{\textbf{User-Centric Personalization}} & \multirow{2}{*}{\textbf{Model}} \\
\cmidrule(lr){2-5} \cmidrule(lr){6-9}
 & ODI & DCA & FTP & \textbf{avg.} & AFM & OSF & SSA & \textbf{avg.} & \textbf{rating}\\
\midrule
Claude Sonnet 4 & 4.70 & 4.75 & 4.54 & 4.66 & 4.73 & 4.75 & 4.76 & 4.75 & 4.71 \\
Claude Sonnet 4.5 & 4.60 & 4.72 & 4.39 & 4.57 & 4.54 & 4.70 & 4.68 & 4.64 & 4.61 \\
Claude Haiku 4.5 & 4.55 & 4.70 & 4.29 & 4.51 & 4.66 & 4.68 & 4.69 & 4.68 & 4.60 \\
DeepSeek-R1 & 4.72 & 4.71 & 4.67 & 4.70 & 4.71 & 4.68 & 4.71 & 4.70 & 4.70 \\
Qwen3-Next & 4.74 & 4.51 & 4.70 & 4.65 & 4.71 & 4.75 & 4.75 & 4.74 & 4.70 \\
Gemini 2.5 Pro & \textbf{4.79} & \textbf{4.79} & \textbf{4.74} & \textbf{4.77} & \textbf{4.77} & \textbf{4.79} & \textbf{4.77} & \textbf{4.78} & \textbf{4.78} \\
Gemini 2.5 Flash & 4.76 & 4.74 & 4.54 & 4.68 & 4.72 & 4.73 & 4.76 & 4.74 & 4.71 \\
\bottomrule
\end{tabular}%
}
\caption{Model performance across all rubrics at iteration $t = 1$. (ODI: Output Delivery Integrity, DCA: Domain
Conceptual Alignment, FTP: Functional Task Progression; AFM: Anticipatory Flow Management, OSF: Output
Structure Fit, SSA: Sustained Style Adherence).}
\label{tab:round1_results}
\end{table*}

\begin{table*}[h!]
\centering
\resizebox{0.9\linewidth}{!}{%
\begin{tabular}{lccccccccc}
\toprule
\multirow{2}{*}{\textbf{Test models}} & \multicolumn{4}{c}{\textbf{Task Completeness}} & \multicolumn{4}{c}{\textbf{User-Centric Personalization}} & \multirow{2}{*}{\textbf{Overall}} \\
\cmidrule(lr){2-5} \cmidrule(lr){6-9}
 & ODI & DCA & FTP & \textbf{avg.} & AFM & OSF & SSA & \textbf{avg.} & \textbf{rating} \\
\midrule
Claude Sonnet 4 & 4.70 & 4.79 & 4.39 & 4.63 & 4.77 & 4.78 & 4.79 & 4.78 & 4.71 \\
Claude Sonnet 4.5 & 4.52 & 4.74 & 3.99 & 4.42 & 4.37 & 4.73 & 4.66 & 4.59 & 4.51 \\
Claude Haiku 4.5 & 4.32 & 4.71 & 3.75 & 4.26 & 4.60 & 4.66 & 4.69 & 4.65 & 4.46 \\
DeepSeek-R1 & 4.72 & 4.71 & 4.63 & 4.69 & 4.71 & 4.66 & 4.71 & 4.69 & 4.69 \\
Qwen3-Next & 4.77 & 4.30 & 4.69 & 4.59 & 4.70 & 4.79 & 4.78 & 4.76 & 4.68 \\
Gemini 2.5 Pro & \textbf{4.88} & \textbf{4.88} & \textbf{4.78} & \textbf{4.85} & \textbf{4.82} & \textbf{4.88} & \textbf{4.84} & \textbf{4.85} & \textbf{4.85} \\
Gemini 2.5 Flash & 4.81 & 4.77 & 4.36 & 4.65 & 4.72 & 4.74 & 4.80 & 4.75 & 4.70 \\
\bottomrule
\end{tabular}%
}
\caption{Model performance across all rubrics at iteration $t = 2$. (ODI: Output Delivery Integrity, DCA: Domain
Conceptual Alignment, FTP: Functional Task Progression; AFM: Anticipatory Flow Management, OSF: Output
Structure Fit, SSA: Sustained Style Adherence).}
\label{tab:round2_results}
\end{table*}

\end{document}